\documentclass[journal]{IEEEtran}

\usepackage{enumerate}
\usepackage{cite}
\usepackage{graphicx}
\usepackage{float}
\usepackage{amsmath}
\newtheorem{Definition}{Definition}

\usepackage{amsfonts} 

\usepackage{algorithmic}
\usepackage{algorithm}

\usepackage{url}
\usepackage[justification=centering]{caption}
\usepackage{color,xcolor}

\begin{document}

\title{Low-latency Federated Learning and Blockchain for Edge Association in Digital Twin empowered 6G Networks}
%
% authors infomation
\author{Yunlong~Lu,~\IEEEmembership{Student Member,~IEEE},
        Xiaohong~Huang,~\IEEEmembership{Member,~IEEE},
        % Yueyue~Dai,~\IEEEmembership{Student Member,~IEEE},
        Ke~Zhang,
        Sabita Maharjan,~\IEEEmembership{Senior~Member,~IEEE},
        and~Yan~Zhang,~\IEEEmembership{Fellow,~IEEE}% <-this % stops a space

\thanks{This work was partially supported by the National Natural Science Foundation of China under Grant No. 61941102, and in part by the Opening Project of Shanghai Trusted Industrial Control Platform, under Grant No. TICPSH202003016-ZC.}
\thanks{Y. Lu, X. Huang are with the Institute of Network Technology, Beijing University of Posts and Telecommunications, Beijing, China, (e-mail: yunlong.lu@ieee.org; huangxh@bupt.edu.cn;).}% <-this % stops a spac
\thanks{K. Zhang is with the School of Information and Communication Engineering, University of Electronic Science and Technology of China, (email:zhangke@uestc.edu.cn).}% <-this % stops a space
\thanks{Sabita Maharjan and Yan Zhang are with Department of Informatics, University of Oslo, Norway; and Simula Metropolitan Center for Digital Engineering, Norway (e-mail: sabita@ifi.uio.no; yanzhang@ieee.org).}% <-this % stops a space
}

\maketitle
\begin{abstract}
Emerging technologies such as digital twins and 6th Generation mobile networks (6G) have accelerated the realization of edge intelligence in Industrial Internet of Things (IIoT). The integration of digital twin and 6G bridges the physical system with digital space and enables robust instant wireless connectivity. With increasing concerns on data privacy, federated learning has been regarded as a promising solution for deploying distributed data processing and learning in wireless networks. However, unreliable communication channels, limited resources, and lack of trust among users, hinder the effective application of federated learning in IIoT. In this paper, we introduce the Digital Twin Wireless Networks (DTWN) by incorporating digital twins into wireless networks, to migrate real-time data processing and computation to the edge plane. Then, we propose a blockchain empowered federated learning framework running in the DTWN for collaborative computing, which improves the reliability and security of the system, and enhances data privacy. Moreover, to balance the learning accuracy and time cost of the proposed scheme, we formulate an optimization problem for edge association by jointly considering digital twin association, training data batch size, and bandwidth allocation.
We exploit multi-agent reinforcement learning to find an optimal solution to the problem. Numerical results on real-world dataset show that the proposed scheme yields improved efficiency and reduced cost compared to benchmark learning method.
\end{abstract}

% Note that keywords are not normally used for peerreview papers.
\IEEEpeerreviewmaketitle

\section{Introduction}
Recent advancements in digital twins and 6th Generation mobile networks (6G) have paved the way for Industrial Internet of Things (IIoT) \cite{8658105}. A series of edge intelligence use cases such as intelligent transportation \cite{9072340,zk3} and smart city \cite{7945258} are envisioned to provide high-quality services to users in the IIoT network. Data processing based on Artificial Intelligence (AI) algorithms lays the foundation for enabling the provision of intelligent services in the IIoT framework. Despite such enabling technologies and recent developments, the gap between the outcome of data analysis and the true reflection of the status of their physical systems remains a major hurdle in designing robust control algorithms for physical systems \cite{uhlemann2017digital}. In this regard, the digital twin paradigm \cite{8477101} emerges as one of the most promising technologies in 6G networks that can enable the near-instant wireless connectivity and extreme-reliable wireless communication\cite{dang2020should}. Empowered by digital twin, the 6G networks can bridge the physical systems with the digital world in real-time for realizing robust edge intelligence in IIoT. 

Digital twins represent real objects or subjects with their data, functions, and communication capabilities in the digital space \cite{8289327}. To reduce latency and enhance reliability for edge computing applications, we incorporate digital twins into wireless networks and propose the Digital Twin Wireless Networks (DTWN). By mapping IoT devices to digital twins in edge servers, DTWN improve the efficiency of AI algorithms and mitigates the impact of unreliable communication between users and edge servers. Moreover, DTWN interact with the physical IIoT systems in real-time through data collection and analytics to keep the digital plane synchronised with the physical system. Thus, the running state analysis and optimization of user devices can be conducted directly in the DTWN, which reduces the system cost. We have noted few works that discuss integrating digital twins with physical systems to improve their performance for different applications. For instance, in \cite{8289327}, the authors proposed the Experimentable Digital Twins (EDTs) that can be used to realize complex control algorithms or mental models.
% in \cite{8289327}, the authors created the digital representations of the respective real assets and their behaviors through simulation techniques to build Experimentable Digital Twins (EDTs), which can be used to realize complex control algorithms or mental models for intelligent systems.
In particular, the authors constructed a general digital twin for complex equipment and proposed a new method for health management using digital twins \cite{tao2018digital}. In this paper, we propose to integrate digital twins into edge networks to mitigate the resource gap between end users and Base Stations (BSs) for efficient edge computing.

Conventional centralized AI algorithms require data to be transmitted to the server, which incurs high data leakage risks. To enhance data security and protect user privacy, federated learning \cite{mcmahan2017communication,smith2017federated} has emerged as a new paradigm for distributed machine learning. Different from conventional AI algorithms, federated learning based models require transmitting only the parameters from locally trained models to the server while the original data is stored with the user to alleviate security and privacy risks \cite{8843900}. To this end, the distributed framework and the enhanced data privacy assurance offered by federated learning have attracted considerable attention and have been well studied for typical MEC applications \cite{9060868} such as content caching \cite{yu2018federated} and vehicular data sharing recently \cite{8998397}. 
% The authors in \cite{yu2018federated} proposed to use federated learning to predict the popularity of different files for content caching in edge computing, which protects content privacy. In \cite{8998397}, the authors proposed to use asynchronous federated learning in the Internet of Vehicles (IoV) \cite{contreras2017internet} for vehicular data sharing. 
Nonetheless the need of conducting the iterative training process on distributed entities each with possibly limited computing resources can adversely affect the efficiency of federated learning in edge networks. To address the issue, in \cite{wang2019adaptive}, the authors proposed a control algorithm to determine the trade-off between local update and global aggregation under a given resource budget. In \cite{8737464}, the authors formulated resource consumption and learning performance into an optimization problem and provided the approximate optimal solution. 

Moreover, the lack of mutual trust among users is another key factor that can hinder them from participating in the learning process to realize edge intelligence. To deal with this issue, blockchain has been widely explored to build a secure collaborative learning mechanism among untrusted users. For example, the authors in \cite{alcaraz2020blockchain} proposed a three layer-based architecture to manage reliable and secure connections among entities, processes and critical resources for the smart grid. Meanwhile, the efficiency and intelligence of blockchain needs to be further improved. Several studies have explored the synergy between AI algorithms and blockchain to enhance the data security and system reliability aspects in edge intelligence enabled networks. In \cite{zk1}, the authors presented an edge intelligence and blockchain empowered framework to enhance the security of edge service management in an IoT scenario. Similarly, in \cite{8998330}, the authors integrated blockchain with Deep Reinforcement Learning (DRL) for content caching in vehicular networks. As conventional AI algorithms can suffer much data transmission load while also leading to higher data leakage risk, we propose to integrate blockchain with federated learning to achieve secure and efficient edge intelligence. In our proposed framework, the federated learning parameters are recorded in the blockchain instead of the parameter server thus considerably enhancing the security of parameters and improving the reliability of models from users.

% We incorporate digital twins into BSs and run our proposed permissioned blockchain empowered federated learning framework on the digital twins instead of end users. 
In our proposed framework, DTWN mitigate the unreliability of learning parameter transmission caused by long communication distance and limited communication resource between end users and edge servers (BSs). However, some new challenges also arise for the maintenance of digital twins. First is the associations of digital twins to various servers. It consumes considerable amount of resources to maintain digital twins by synchronizing real-time data and building corresponding models. Due to the dynamic computing and communication resources available with edge servers, the association of digital twins to corresponding servers is a crucial factor in determining the performance of DTWN, which is a largely unexplored dimension in the context of edge networks. Moreover, due to the multiple communication rounds in federated learning, the limited bandwidth available with BSs requires to be optimally allocated to improve the communication efficiency of associated edge servers. In this paper, we first design the new DTWN model by incorporating digital twins into wireless networks to mitigate the unreliable and long distance communication between users and BSs, where the user data is synchronized to BSs to construct corresponding digital twins. Based on the DTWN, we propose a permissioned blockchain empowered federated learning framework for realizing robust edge intelligence. 
% Analysis of data and behaviors of end users 
Data analysis can be executed on their digital twins in the BS plane directly instead of in the resource constrained end users. The security and privacy of digital twin data are also enhanced by . To improve the efficiency of the proposed architecture, we formulate an optimization problem for edge association and bandwidth allocation, and design a multi-agent reinforcement learning based algorithm to find the optimal solution to it. The main contributions of this paper are summarized as follows.
\begin{itemize}
	\item We design a new digital twin wireless network model- DTWN, which uses digital twins to mitigate the unreliable and long distance communication between end users and edge servers. 
	\item We define the edge association problem for DTWN, and propose a permissioned blockchain empowered federated learning framework for edge computing in DTWN.
	\item To improve the efficiency of the proposed scheme, we formulate an optimization problem by jointly considering edge association and communication resource allocation to minimize the time cost, and develop a multi-agent reinforcement learning based algorithm to find the optimal solution to the problem.
\end{itemize}

% The rest of the paper is organized as follows. In Section \ref{section:model}, the system model of DTWN is provided, in which we develop the blockchain and federated learning framework running on digital twins. In Section \ref{section:problem}, we formulate a new edge association problem in DTWN to minimize the total system latency. A multi-agent reinforcement learning based algorithm is developed to find the optimal solution to the problem in Section \ref{section:solution_DRL}. Numerical results are presented in Section \ref{section:evaluation}. Finally, Section \ref{section:conclusion} concludes the paper.

The rest of the paper is organized as follows. In Section \ref{section:model}, the blockchain and federated learning model for DTWN is provided. In Section \ref{section:problem}, we formulate a new edge association problem in DTWN to minimize the total system latency. A multi-agent reinforcement learning based algorithm is developed to find the optimal solution to the problem in Section \ref{section:solution_DRL}. Numerical results are presented in Section \ref{section:evaluation}. Finally, Section \ref{section:conclusion} concludes the paper.
% However, the performance of the above work greatly relies on choosing the correct threshold, which is not an easy task. A "perfectly" fixed threshold can solve the problem of one specific scenario, but for another problem scenario it may be a disaster. Moreover, the threshold selection requires a lot of time to debug, which is unrealistic for scenarios requiring high-speed response. At this time, the Lazily Aggregated Gradient (LAG) [28]

\section{Blockchain and Federated Learning Model for Digital Twin Networks}
\label{section:model}
In this section, we first present our DTWN model. Then we describe the edge association problem in DTWN in Section \ref{section:model_ass}, which is followed by the federated learning model in Section \ref{section:model_FL}. Then we introduce our blockchain model in Section \ref{section:model_blockchain} and the wireless communication model in Section \ref{section:model_comm}.

\begin{figure}[htb]
	\centering
	\includegraphics[scale=0.23]{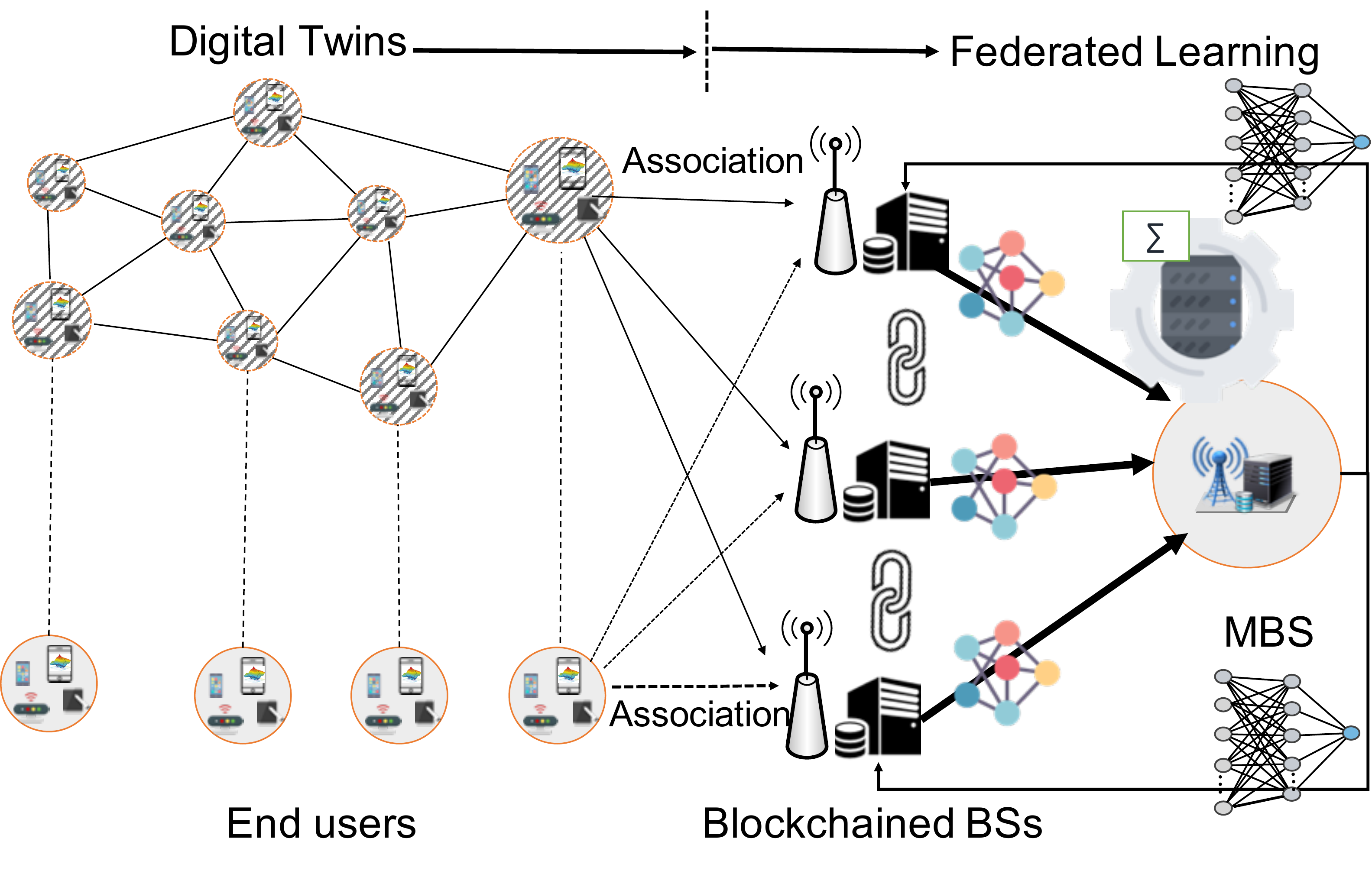}
	\caption{The proposed Digital Twin Wireless Networks}
	\label{fig:DTWN}
\end{figure}

We consider a blockchain and federated learning empowered digital twin network model as depicted in Fig. \ref{fig:DTWN}. The system consists of $N$ end users such as IoT devices and mobile devices, $M$ BSs, and a Macro Base Station (MBS). The BSs and the MBS are equipped with MEC servers. The end devices generate running data and synchronize their data with corresponding digital twins that run on the BSs. We use $\mathcal{D}_i = \{ (x_{i1},y_{i1}),...,(x_{iD_i},y_{iD_i})\}$ to denote the data of end user $i$, where $D_i$ is the data size, $x_{i}$ is the data collected by end users and $y_i$ is the label of $x_{i}$. The digital twin of end user $i$ in the BSs are denoted as $DT_i$, which is composed of behavior model  $\mathcal{M}_i$, static running data $\mathcal{D}_i$ and real-time dynamic state $s_t$, $DT_i = (\mathcal{M}_i,\mathcal{D}_i,s_t)$. $\mathcal{D}_i$ and $s_t$ is the essential data required to run digital twin applications. Instead of synchronizing all raw data to digital twins, which incurs great communication load and data leakage risk, we use federated learning to learn model $\mathcal{M}$ from user data. In various application scenarios, the end users may communicate with other end users to exchange running information and share data, through e.g., Device to Device (D2D) communications. Thus, the digital twins also form a network based on the connections of end users. Based on the constructed DTWN, we can obtain the running states of physical devices and make further decisions to optimize and drive the running of devices by analyzing digital twins directly.

In our proposed digital twin network model, we use federated learning to execute the training and learning process collaboratively for edge intelligence. Moreover, since the end users lack mutual trust and the digital twins consist of private data, we use permissioned blockchain to enhance the system security and data privacy. The permissioned blockchain records the data from digital twins and manages the participating users through permission control. The blockchain is maintained by the BSs, which are also the clients of the federated learning model. The MBS runs as the server for the federated learning model. In each iteration of federated learning, the MBS distributes the machine learning model parameters to BSs for training. The BSs train the model based on data from digital twins and returns the model parameters to the MBS.
\subsection{Edge Association in DTWN}
\label{section:model_ass}
The end devices or users are mapped to the digital twins in BSs in DTWN. The maintenance of digital twins consumes a large amount of computing and communication resources for synchronizing real-time data and building corresponding models. However, the computation and communication resources in wireless networks are very limited, which should be optimally used to improve the resource utility. Thus, to associate various IoT devices with different BSs according to their computation capabilities and states of the communication channel, is a key problem in DTWN. As depicted in Fig. \ref{fig:DTWN},  The digital twins of IoT devices are constructed and maintained by their associated BSs. The training data and the computation tasks for training are distributed to various BSs based on the association between digital twins and the BSs. We provide the formal definition of the edge association in Definition \ref{Definition:EA}.

\begin{Definition}[Edge Association] 
\label{Definition:EA}
Consider a DTWN with $N$ IoT users and $M$ BSs. For any user $u_i$, $i \in \mathcal{N}$, the goal of edge association is to choose the target BS $j \in \mathcal{M}$ to construct the digital twin $DT_i$ of user $i$. The association $\langle DT_i, BS_j\rangle$ is denoted as $\Phi(i,j)$. If $DT_i$ is associated to BS $j$, then $\Phi(i,j)= D_i$, where $D_i$ is the size of data used to construct $DT_i$. Otherwise, $\Phi(i,j)= 0$.
\end{Definition}

A BS can associate with multiple digital twins, while a digital twin can only be associated with at most one BS. That is, $\sum_{j=1}^{M}\Phi(i,j) = D_i$. We perform edge association according to the datasets $D_i$ of IoT users, the computation capability of BSs $f_j$, and the transmission rate $R_{i,j}$ between $u_i$ and $BS_j$, denoted as
\begin{equation}
\label{eq:ass}
\Phi(i,j) = f(D_i,f_j,R_{i,j}).
\end{equation}
The objective of the edge association problem is to improve the utility of resources and the efficiency of running digital twins in the DTWN.

\subsection{Federated Learning Model}
\label{section:model_FL}

The goal of federated learning is to train a global machine learning model $\mathcal{M}$ based on the data from various digital twins $DT_i$ without transmitting the original training data. The learning process involves finding the parameter $w$ for global model $\mathcal{M}$ that minimizes the global loss function $f(w,x,y)$, denoted as 

\begin{equation}
\label{eq:loss}
\min_w \frac{1}{N}\sum_{i=1}^{N}\frac{1}{D_j}\sum_{j=1}^{D_j}f(w,x_{ij},y_{ij}).
\end{equation}

The client BSs train their local models based on their own data. Each client minimizes the loss function on its local data through gradient descent based methods such as gradient descent and stochastic gradient descent, with a predefined learning rate. The clients then transmit the local gradients or the local models to the server MBS to update the global model, through algorithms such as federated averaging (FedAvg) \cite{mcmahan2017communication} and Federated SGD (FedSGD). We denote the aggregation of the global model as:

\begin{equation}
\label{eq:aggregation}
\mathcal{M} = \frac{1}{N}\sum_{i=1}^{N}D_i w_i.
\end{equation}
In the training process, the BSs train the local models based on the digital twins $DT_i$ that they maintained according to Eq. \ref{eq:loss}. A BS may contain several digital twins constructed from the end users under its coverage. Different from conventional federated learning, the BSs in our proposed scheme first aggregate their local models from multiple digital twins instead of sending the original local models to the MBS, which can reduce the transmission load. As BS $m$ has $K_i$ digital twins, the aggregation at BS $i$ is

\begin{equation}
\label{eq:BS_agg}
\mathcal{G}_m = \frac{1}{K_i}\sum_{j=1}^{K_i}D_{DT_j} w_{DT_j},
\end{equation}
where $D_{DT_i}$ is the training data of digital twin $DT_i$ and $w_{DT_i}$ is the trained model of $DT_i$. BS $m$ then sends $\mathcal{G}_m$ to the MBS, and the MBS updates the global model as
\begin{equation}
\label{eq:MBS_agg}
\mathcal{M} = \frac{1}{M}\sum_{m=1}^{M}\mathcal{G}_m.
\end{equation}
The local models of digital twins are transmitted to the MBS through wireless links. The MBS collects the model parameters from all participating BSs and updates the global model. In this process, since all BSs transmit their parameters in the same period and the wireless bandwidth is limited, the communication efficiency is of vital importance for the learning phase to converge.
\subsection{Blockchain Model}
\label{section:model_blockchain}
To enhance the security and reliability of digital twins from untrusted end users, the BSs act as the blockchain nodes and maintain the running of the permissioned blockchain. The digital twins are stored in the blockchain and their data is updated as the state of the corresponding user changes. Moreover, the local models of the BS are also stored in the blockchain, which can be verified by other BSs to ensure the quality of local models. Thus, there are three types of records, namely, digital twin model records, digital twin data records, and training model records.

\begin{figure}[htb]
	\centering
	\includegraphics[scale=0.28]{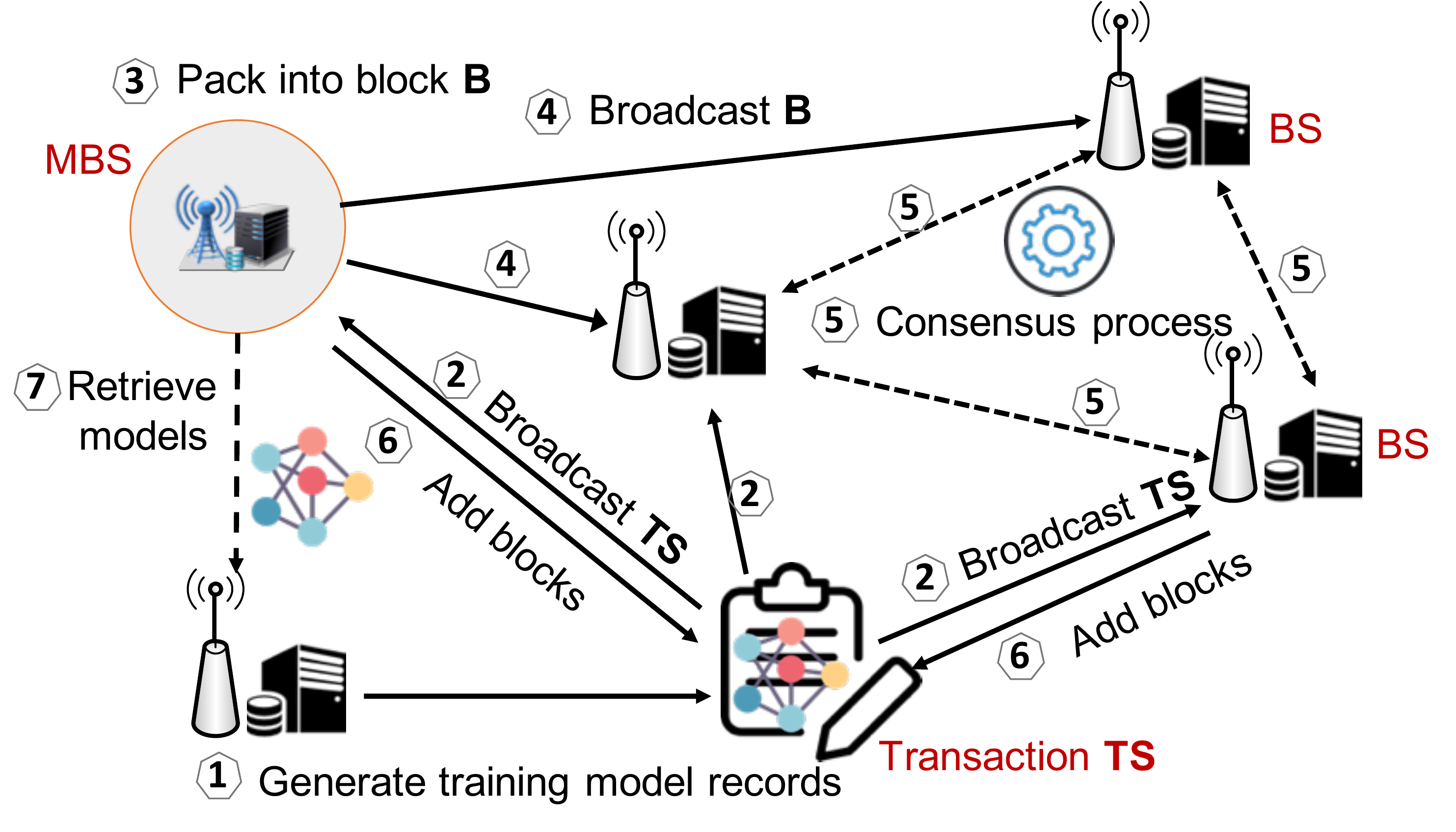}
	\caption{The blockchain scheme for federated learning}
	\label{fig:blockchain}
\end{figure}

The overall blockchain scheme for federated learning is shown in Fig. \ref{fig:blockchain}. The BSs first train the local training models on their own data, and then upload the trained models to the MBS. The trained models are also recorded as blockchain transactions, and broadcasted to other BSs for verification. Other BSs collect the transactions and pack them into blocks. The consensus process is executed to verify the transactions in blocks.
Our consensus process is executed based on Delegated Proof of Stakes (DPoS) protocol, 
% as shown in Fig. \ref{fig:consensus}. In our consensus,b
where the stakes are the training coins. Initial training coins are allocated to BS $i$ according to its data from digital twins, denoted as
\begin{equation}
\label{eq:stake}
S_{i} = \frac{\sum_{j=1}^{K_i} D_{DT_j}}{\sum_{k=1}^{M}D_{k}}S_{ini},
\end{equation}
where $S_{ini}$ is an initial value, $K_i$ is the number of digital twins associated to BS $i$.
% \begin{figure}[htb]
% 	\centering
% 	\includegraphics[scale=0.25]{figures/consensus.pdf}
% 	\caption{The consensus process of the permissioned blockchain}
% 	\label{fig:consensus}
% \end{figure}

The coins of each BS are then adjusted according to their performance in the training process. If the trained model of a BS passes the verification of other BSs and the MBS, coins will be awarded to the BS. Otherwise, the BS will get no pay for its training work. A number of BSs are elected as the block producers by all BSs. In the voting process, all the BSs vote for the candidate BSs by using their own training coins. The elected BSs take turns to pack the transactions in a time interval $T$ into a block $B$, and broadcast block $B$ to other producers for verification. 

In our federated learning scheme, we leverage blockchain to verify the local models before embedding them into the global model. Due to high resource consuming for block verification,
the interval $T$ should be set to multiple times of the local training period, that is, the BSs execute multiple local training iterations before transmitting local models to the MBS for global aggregation. 
\subsection{Communication Model}
\label{section:model_comm}
We use the Orthogonal Frequency Division Multiple Access (OFDMA) for wireless transmission in our system. To upload trained local models, all BSs share $C$ sub-channels to transmit their parameters. The achievable uplink data rate from BS $i$ to the MBS is: 
\begin{equation}
\label{eq:uplink}
R_i^U = \sum_{c=1}^{C} \tau_{i,c} W^U log_2 (1+\frac{P_{i,c}^U h_{i,c}^U r_{i,m}^{- \alpha}}{\sum_{j \in \mathcal{N}'} P_{j,c}^U h_{j,c}^U r_{i,m}^{- \alpha} + N_0}),
\end{equation}
where $C$ is the total number of subchannels, $\tau_{i,c}$ is the time fraction allocated to BS $i$ on subchannel $c$, $W^U$ is the bandwidth of each subchannel which is a constant value. The transmission power is $P_{i,c}^U$ and the uplink channel gain on subchannel $c$ is $h_{i,c}^U$. $r_{i,m}^{- \alpha}$ is the path loss fading of the channel between BS $i$ and the MBS, $r_{i,m}$ is the distance between BS $i$ and the MBS, and $\alpha$ is the path loss exponent. $N_0$ is the noise power and $\sum_{j \in \mathcal{N}'} P_{j,c}^U h_{j,c}^U r_{i,m}^{- \alpha}$ is the interference caused by other BSs using the same subchannel.
In the download phase, the MBS broadcasts the global model with the rate 
\begin{equation}
\label{eq:downlink}
R_i^D = \sum_{c=1}^{C}  W^D log_2 (1+\frac{P_{i,c}^D h_{i,c}^D}{\sum_{j \in \mathcal{N}''} P_{j,c}^D h_{j,c}^D r_{i,m}^{- \alpha} +N_0}),
\end{equation}
where $P_{i,c}^D$ is the downlink power of BS $i$ and $h_{i,c}^D$ is the channel gain between BS $i$ and the MBS, $\sum_{j \in \mathcal{N}''} P_{j,c}^D h_{j,c}^D r_{i,m}^{- \alpha}$ is  the downlink inference. 

\section{Edge Association in DTWN: Problem Formulation}
\label{section:problem}
In this section, we first analyze the delay performance of the proposed model, and then formulates the optimization problem for edge association in DTWN to minimize the total system time cost under the given learning accuracy requirement.

Now we start to derive the formulation of edge association problem. We consider that the gradient $\nabla f(w)$ of $f(w)$ is $L$-Lipschitz smooth, that is 
\begin{equation}
\label{eq:L}
||\nabla f(w_{t+1})-\nabla f(w_t)|| \leq L ||w_{t+1}-w_t||,
\end{equation}
where $L$ is a positive constant and $||w_{t+1}-w_t||$ is the norm of $w_{t+1}-w_t$. We consider that the loss function $f(w)$ is strongly convex, that is
\begin{equation}
\label{eq:c}
f(w_{t+1}) \geq f(w_t) + \langle \nabla f(w_t),w_{t+1}-w_t \rangle + \frac{1}{2}||w_{t+1}-w_t||^2.
\end{equation}
Many loss functions for federated learning can satisfy the above assumptions, for example, the logic loss functions. If (\ref{eq:L}) and (\ref{eq:c}) are satisfied, the upper bound of global iterations can be obtained as \cite{ma2017distributed}
\begin{equation}
\label{eq:bound}
\mathcal{T}(\theta_L,\theta_G) = \frac{\mathcal{O}(log(1/\theta_L))}{1-\theta_G},
\end{equation}
where $\theta_L$ is the local accuracy $\frac{||\nabla f(w_{t+1})||}{||\nabla f(w_t)||} \leq \theta_L$, $\theta_G$ is the global accuracy, and $0  \leq \theta_L, \theta_G\leq 1$. As in \cite{8737464}, we consider $\theta_L$ as a fixed value, so that the upper bound $\mathcal{T}(\theta_L,\theta_G)$ can be simplified to $\mathcal{T}(\theta_G) = \frac{1}{1-\theta_G}
$.
Denote the time of one local training iteration by $T_{cmp}$, then the computation time in one global iteration is $log(1/\theta)T_{cmp}$, and the upper bound of total learning time is $\mathcal{T}(\theta_G)T_{glob}$.

The time cost in our proposed scheme mainly consists of 

\begin{enumerate}
	\item \textit{Local training on digital twins:} The time cost for the local training of BS $i$ is decided by the computing capability and the data size of its digital twins. The time cost is 
		\begin{equation}
		\label{eq:local_training}
		T_i^{cmp} = \frac{\sum_{j=1}^{K_i} b_j D_{DT_j}}{f_i^C}f^C,
		\end{equation}
		where $f^C$ is the number of CPU cycles required to train one sample of data, ${f_i^C}$ is the CPU frequency of BS $i$, $b_j$ is the training batchsize of digital twin $DT_j$.
	\item \textit{Model aggregation on BSs:} The BSs aggregate its local models from various digital twins according to Eq. (\ref{eq:BS_agg}). The computing time for local aggregation is
		\begin{equation}
		\label{eq:local_agg}
		T_i^{la} = \frac{\sum_{j=1}^{K_i} |w_j|}{f_i^C}f_b^C,
		\end{equation}
		where $|w_j|$ is the size of local models and $f_b^C$ is the number of CPU cycles required to aggregate one unit of data. Since all the clients share the same global model, $|w_1|=|w_2|=...=|w_j|=|w_g|$. Thus the time cost for local aggregation is 
		\begin{equation}
		\label{eq:local_agg2}
		T_i^{la} = \frac{K_i |w_g|}{f_i^C}f_b^C.
		\end{equation}
	\item \textit{Transmission of model parameters:}
		The local models aggregated by BS $i$ are then broadcasted to other BSs as transactions. The time cost is related to the number of blockchain nodes and the transmission efficiency. Since other BSs also help to transmit the transaction in the broadcast process, the time function is related to $log_2 M$, where $M$ is the size of the BS network. The required time cost is 
		\begin{equation}
		\label{eq:bs_broadcast}
		T_i^{pt} = \xi log_2 M \frac{K_i |w_g|}{R_i^U},
		\end{equation}
		where $\xi$ is a factor of transmission time cost that can be obtained from historical running records of the transmission process.
	\item \textit{Block Validation:}
		The block producer BS collects the transactions and packs them into a block. The block is then broadcasted to other producer BSs, and validated by them. Thus, the time cost is 
		\begin{equation}
		\label{eq:block}
		T_{bp}^{bv} = \xi log_2 M_p \frac{S_B}{R_i^D} + \max_i \frac{S_B f^v}{f_is},
		\end{equation}
		where $M_p$ is the number of block producers, $S_B$ is the size of a block.
\end{enumerate}

Note that in the aggregation phase, the size of model parameters $|w_g|$ is small and the computing capability $f_i$ is high. Thus, compared to other phases, the time for aggregation is very small that can be neglected. Based on the above analysis, the time cost for one iteration is denoted as
\begin{equation}
\begin{aligned}
\label{eq:time}
T &= \max_i\left\{\frac{\sum_{j=1}^{K_i} b_j D_{DT_j}}{f_i^C}f^C \right \} + \max_i\left\{ \xi log_2 M \frac{K_i |w_g|}{R_i^U}\right \} \\ & +  \xi log_2 M_p \frac{S_B}{R_i^D} + \max_i \frac{S_B f^v}{f_is},
\end{aligned}
\end{equation}

In the 6G network, the growth of the user scale, the ultra-low latency requirement of communication, and the dynamic network status make it an important issue to reduce the time cost of model training in various applications. Since accuracy and latency are two main metrics to evaluate the decision making abilities of digital twins in our proposed scheme, we consider the edge association problem to find the trade-off between learning accuracy and  time cost of the learning process. Due to the dynamic computing and communication capabilities of various BSs, the edge association of digital twins, that is, how to allocate the digital twins of different end users to various BSs for training, is a key issue to be solved for minimizing the total time cost. Moreover, increasing the training batch size $b_n$ of each digital twin $DT_n$ can improve the learning accuracy. However, it will also increase the learning time cost to execute more computation. In addition, how to allocate the bandwidth resources to improve the communication efficiency, should also be considered. In our edge association problem, we should carefully design these policies to minimize the total time cost of the proposed scheme. Thus, we formulate the optimization problem as minimizing the time cost of federated learning under a given learning accuracy. To solve the problem, the association of digital twins, the batchsize of their training data, the bandwidth allocation should be jointly considered according to the computing capability $f_i^C$ and the channel state $h_{i,c}$. The optimization problem can be formulated as
\begin{align}
 & \mathop{\min} \limits_{\boldsymbol{K_i},\boldsymbol{b_n},\boldsymbol{\tau_{i,c}}} \frac{1}{1-\theta_G}T \label{eq:problem}\\
 \mbox{s.t.} \quad & \theta_G \geq \theta_{th}, \theta_G, \theta_{th} \in (0,1), \tag{\ref{eq:problem}{a}} \label{eq:problem_a}\\
 & \sum_{i=1}^{M} K_{i} = D,  K_i \in \mathcal{N}, \tag{\ref{eq:problem}{b}} \label{eq:problem_b}\\
 & \sum_{i=1}^{M} \tau_{i,c} \leq 1, c \in \mathcal{C}, \tag{\ref{eq:problem}{c}} \label{eq:problem_c}\\
 & b^{min} \leq b_n \leq b_n^{max}, \forall n \in \mathcal{N} \tag{\ref{eq:problem}{d}} \label{eq:problem_d}
\end{align}
Constraint (\ref{eq:problem}{b}) ensures that the sum of the number of associated digital twins does not exceed the size of total dataset. Constraint (\ref{eq:problem}{c}) guarantees that each subchannel can only be allocated to at most one BS. Constraint (\ref{eq:problem}{d}) ensures the range of training batch size for each digital twin. Problem (\ref{eq:problem}) is a combinational problem. Since there are several products of variables in the objective function, and the time cost of each BS is also affected by the resource states of other BSs, it is challenging to solve problem (\ref{eq:problem}).

\section{Multi-agent DRL for Edge Association}
\label{section:solution_DRL}
Since the system states are only determined by network states in the current iteration and the allocation policies in the last iteration, we regard the problem as a Markov Decision Process (MDP) and use multi-agent DRL \cite{lowe2017multi} based algorithm to solve it.
\subsection{Multi-agent DRL Framework}
\begin{figure}[htb]
	\centering
	\includegraphics[scale=0.242]{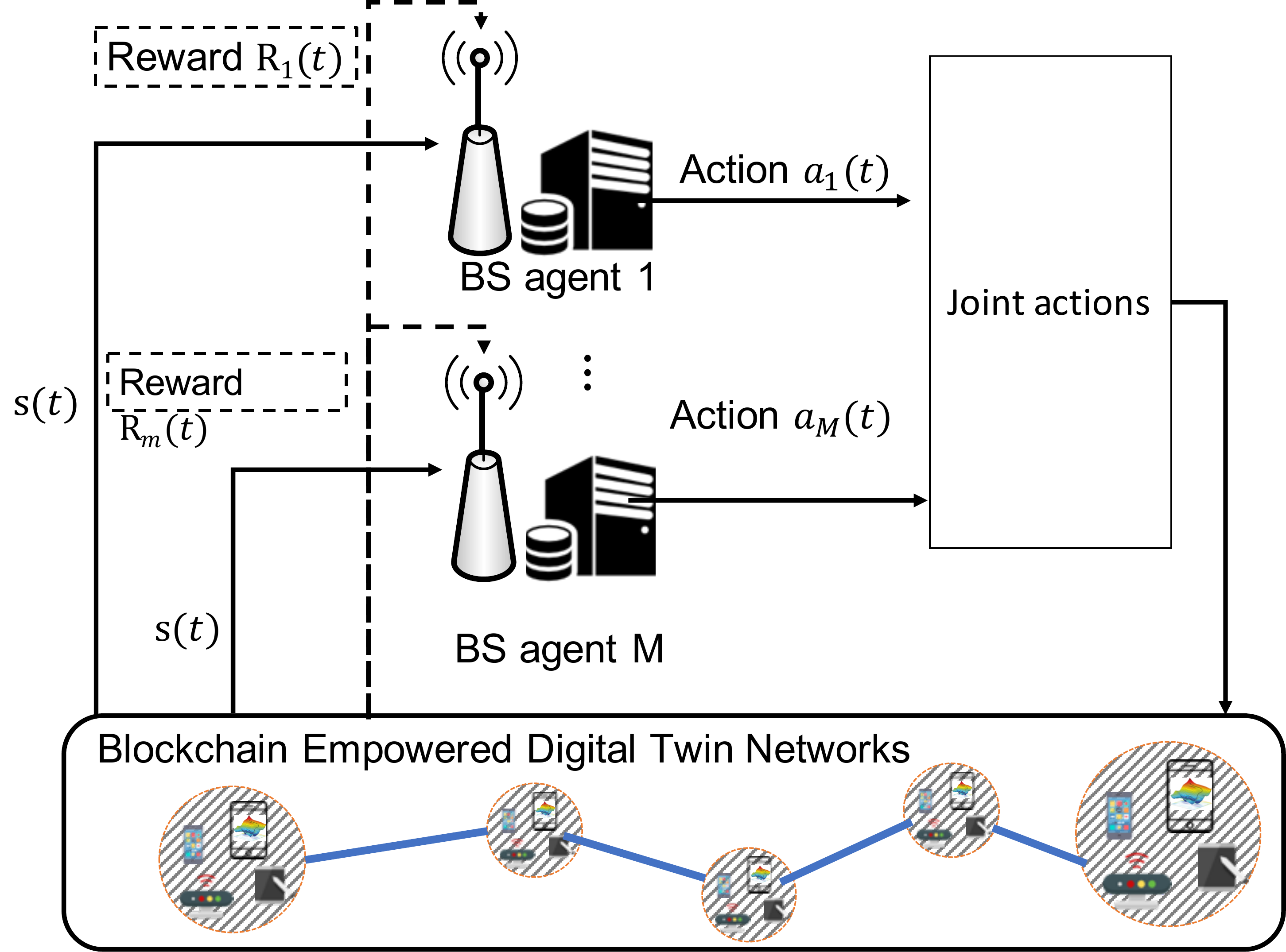}
	\caption{Multi-agent DRL for edge association}
	\label{fig:drl}
\end{figure}

The proposed multi-agent reinforcement learning framework is shown in Fig. \ref{fig:drl}. In our proposed system, each BS is regarded as a DRL agent. The environment consists of BSs and digital twins of end users. Our multi-agent DRL framework consists of multiple agents, a common environment, system state $\mathcal{S}$, action $\mathcal{A}$ and the reward function $\mathcal{R}$, which are described below. 
\begin{itemize}
	\item \textit{State space:} The state of the environment is composed of the computing capabilities $f^C$ of BSs, the number of digital twins $K_i$ on each BS $i$, the training data size of each digital twin $D_n$, the channel state $h_{i,c}$. The states of multiple agents are denoted as $s(t)=(\boldsymbol{f}^C,\boldsymbol{K},\boldsymbol{D},\boldsymbol{h})$, where each dimension is a state vector that contains the states for all agents.
	\item \textit{Action space:} The actions of BS $i$ in our system consist of the digital twin allocation $K_i$, the training data batchsizes for its digital twins $\boldsymbol{b}_i$, and the bandwidth allocation $\boldsymbol{\tau}_i$. Thus, the actions are denoted as $\boldsymbol{a}_i(t) = (K_i, \boldsymbol{b}_i, \boldsymbol{\tau}_i)$. BS agent $i$ makes new action decisions $\boldsymbol{a}_i(t)$ at the beginning of iteration $t$ based on system state $s(t)$. The system action is $\boldsymbol{a}(t) = (\boldsymbol{a}_1,...,\boldsymbol{a}_i,...,\boldsymbol{a}_m)$.
	\item \textit{Reward:} We define the reward function of BS $i$ according to its time cost $T_i$ based on Eq. (\ref{eq:time}), as 
	\begin{equation}
	\label{eq:reward}
	\mathcal{R}_i(s(t),\boldsymbol{a}_i(t))= - T_i(t),
	\end{equation}
	The reward vector of all agents is $\boldsymbol{R}= (\mathcal{R}_1,...,\mathcal{R}_m)$. According to Eq. (\ref{eq:problem}), the total time cost $T$ is decided by the maximum time cost of agents $\max\{T_1,T_2,...,T_m\}$. Thus each DRL agent shares the same reward function in our scheme. In the training process, the BS agents adjust their actions to maximize the reward function, that is, to minimize the system time cost in each iteration.
\end{itemize}
The learning process of BS $i$ is to find the best policy that maps its states to its actions, denoted as $\boldsymbol{a}_i = \pi_i (s)$, where $\boldsymbol{a}_i$ is the action to be taken by BS $i$ for the whole system state $s$. The objective is to maximize the expected reward, that is
\begin{equation}
	\label{eq:total_reward}
	\mathcal{R}_t= \sum_{i}\gamma \mathcal{R}_i(s(t),\boldsymbol{a}_i(t)),
\end{equation}
where $\gamma$ is the discount rate, $0 \leq \gamma \leq 1$. In the conventional DRL framework, it is hard for an agent to obtain the states of others. In our DTWN, the states of digital twins and BSs are recorded in the blockchain. A BS can retrieve records from blockchain to obtain the system states and actions of other agents in the training process.

\subsection{Multi-agent DRL algorithm for edge association}

Since the network states variable such as $\boldsymbol{h}$ and $\boldsymbol{f}^C$ are continuous values, the state and action spaces are large. We adopt Deep Deterministic Policy Gradient (DDPG) algorithm to address the formulated MDP problem. The DDPG algorithm uses policy-based Actor-Critic (AC) to select and evaluate the actions, and uses Deep Q Networks (DQNs) to approximate value functions in actor and critic networks. The multiple agents share the same reward function and train their policy network cooperatively to minimize the time cost.

In the training phase, each BS agent generates the training sample $\langle s,\boldsymbol{a}_i, R_i, s' \rangle$, where $s'$ is the state of DTWN in the next time slot. We leverage DQN with the tuples randomly sampled from replay memory to train the actor network $\pi(s_t|\theta \pi)$ and critic network $Q(s_t,\boldsymbol{a}_i|\theta_Q)$, with parameters $\theta_{\pi}$ and $\theta_{Q}$. There are two sets of actor and critic networks- the primary networks and the target networks. The target networks have the same structure as the primary networks and are used to generate target values for training primary networks. 

Each agent $i$ has a actor DNN $\pi(s_t|\theta_{\pi})$ to determine its action $\boldsymbol{a}_i$ based on current state $s_t$, as
\begin{equation}
\label{Eq:actor}
a_i(t) = \pi_i(s_t|\theta_{\pi_i}) + \mathfrak{N},
\end{equation}
where $\mathfrak{N}$ is the random noise generated by the Ornstein- 
Uhlenbeck process \cite{barndorff2001non}, $\theta_{\pi_i}$ is the explored edge association policy.
The parameters of the primary actor DNN are updated as
\begin{equation}
\label{Eq:actorUpdate}
\theta_\pi = \theta_\pi + \alpha_\pi \cdot \mathbb{E}[\nabla_{a_i} Q(\boldsymbol{s_t},\boldsymbol{a}_1,...,\boldsymbol{a}_i| \theta_Q)|_{\boldsymbol{a_i} = \pi(\boldsymbol{s_t}|\theta_\pi)}\cdot \nabla_{\theta_\pi}\pi(\boldsymbol{s_t})],
\end{equation}
where $\alpha_\pi$ is the learning rate of the actor DNN.
We train the primary critic network of agent $i$ by taking steps in the opposite gradient direction of the loss function as
\begin{equation}
\label{Eq:criticUpdate}
\theta_{Q_i} = \theta_{Q_i} + \alpha_{Q_i} \cdot \mathbb{E}[2(y_t-Q(\boldsymbol{s_t},\boldsymbol{a_i}|\theta_{Q_i}))\cdot \nabla Q(\boldsymbol{s}_t,\boldsymbol{a}_1,...,\boldsymbol{a}_i)],
\end{equation}
where $\alpha_{Q_i}$ is the learning rate of the primary critic DNN, and $y_t$ is the target value generated by target networks, $(\boldsymbol{a}_1,...,\boldsymbol{a}_i)$ are the actions of all DRL agents.

The target networks represent the old version of the primary networks. The parameters of the target action DNN $\theta_{\pi}^T$ and the target critic DNN $\theta_{Q}^T$ are trained in a different way. Agent $i$ updates its target action DNN $\theta_{\pi_i}^T$ and critic DNN $\theta_{Q_i}^T$ as
\begin{align}
\label{Eq:tUpdate}
&\theta_{\pi_i}^T = \beta \theta_{\pi_i} + (1-\beta)\theta_{\pi_i},\\
&\theta_{Q_i}^T = \beta \theta_{Q_i} + (1-\beta)\theta_{Q_i},
\end{align}
where $\beta$ is the update rate.

The proposed multi-agent reinforcement learning algorithm for edge association problem is summarized as Algorithm \ref{algorithm:blockchain}. Each agent first initializes its primary and target critic-actor DNNs, and initializes the edge association policy. Then, the primary actor DNN of agent $i$ generates action $\boldsymbol{a}_i$ based on current state and policy according to Eq. (\ref{Eq:actor}). The observed reward $R_i(e)$ and next state $s_{t+1}$ are calculated, and the tuple $(s_t,a,R_i(e),s_{t+1})$ is stored into the replay memory as training samples. The primary actor DNN and critic DNN are then updated based on Eq. (\ref{Eq:actorUpdate}) and Eq. (\ref{Eq:criticUpdate}). The target networks are updated according to Eq. (\ref{Eq:tUpdate}).
\renewcommand{\algorithmicrequire}{\textbf{Input:}}
\renewcommand{\algorithmicensure}{\textbf{Output:}}
\begin{algorithm}[ht]
\caption{The multi-agent deep reinforcement learning based algorithm for edge association}
\label{algorithm:blockchain}
\begin{algorithmic}[1]
\FOR{each BS $i \leq M$}
	\STATE Initialize primary and target actor-critic networks; Initialize replay memory;
\ENDFOR
\FOR{each BS $i \leq M$}
	\FOR{episode $e \leq E$}
		\STATE Initialize DTWN environment setup;
		\FOR{each time slot $t$}
			\STATE Observe current state $s$ and execute action $\boldsymbol{a}_i$ according to Eq. (\ref{Eq:actor});
			\STATE Compute reward $R_i(t)$, and transit to the next state $s_{t+1}$ based on actions $(\boldsymbol{a}_1,\boldsymbol{a}_2,...,\boldsymbol{a}_n)$ 
			\STATE Store $(s_t,\boldsymbol{a}_i(t),R_i(t),s_{t+1})$ into replay memory
			\STATE Update primary actor and critic networks according to Eq. (\ref{Eq:actorUpdate}) and Eq. (\ref{Eq:criticUpdate})
			\STATE Update target networks according to Eq. (\ref{Eq:tUpdate})
		\ENDFOR
	\ENDFOR
\ENDFOR
\end{algorithmic}
\end{algorithm}

The computation cost training process of the DRL can be performed offline for large number of episodes based on the dynamic system states. The trained DRL models are deployed online to optimally allocate the resources to minimize the time cost of federated learning under given learning accuracy. The actions in our multi-agent DRL for edge association are composed of digital twin association policies $\boldsymbol{K}$, batchsize of training data $\boldsymbol{b}$, and the bandwidth allocation policies $\boldsymbol{\tau}$. Due to the large action and state space, 
% is very large, e.g., there are $M^N$ digital twin association policies alone for $N$ users and $M$ BSs, and the $\boldsymbol{b}$ and $$. 
% The training data batchsize of each digital twin is a continuous variable between $b^{min}$ and $b^{max}$. The allocation policies of bandwidth in OFDMA are also high-dimensional due to the dynamic time fraction values. 
we adopt the actor DNN to generate actions. The complexity of our proposed learning algorithm mainly lies in the training process of actor DNN $\theta_{\pi}$ and critic DNN $\theta_{Q}$. Consider the DNN in our algorithm has $L$ hidden layers, and the average number of neurons in each layer is $L_a$. Thus the time complexity for training a DNN is
% $O(\sum_{i=1}^{L}L_i \cdot L_{i+1})$, which can be simplified to 
$O(L_a^2)$. The total complexity for training the multi-agent DRL model with $M$ agents is $O((M \cdot L_a^2) \cdot E)$. Since the training at each agent is executed in parallel, the time complexity is $O(L_a^2 \cdot E)$, which is related to the DNN size and the number of episodes.

\section{Numerical Results}
\label{section:evaluation}
\begin{figure}[htb]
	\centering
	\includegraphics[scale=0.283]{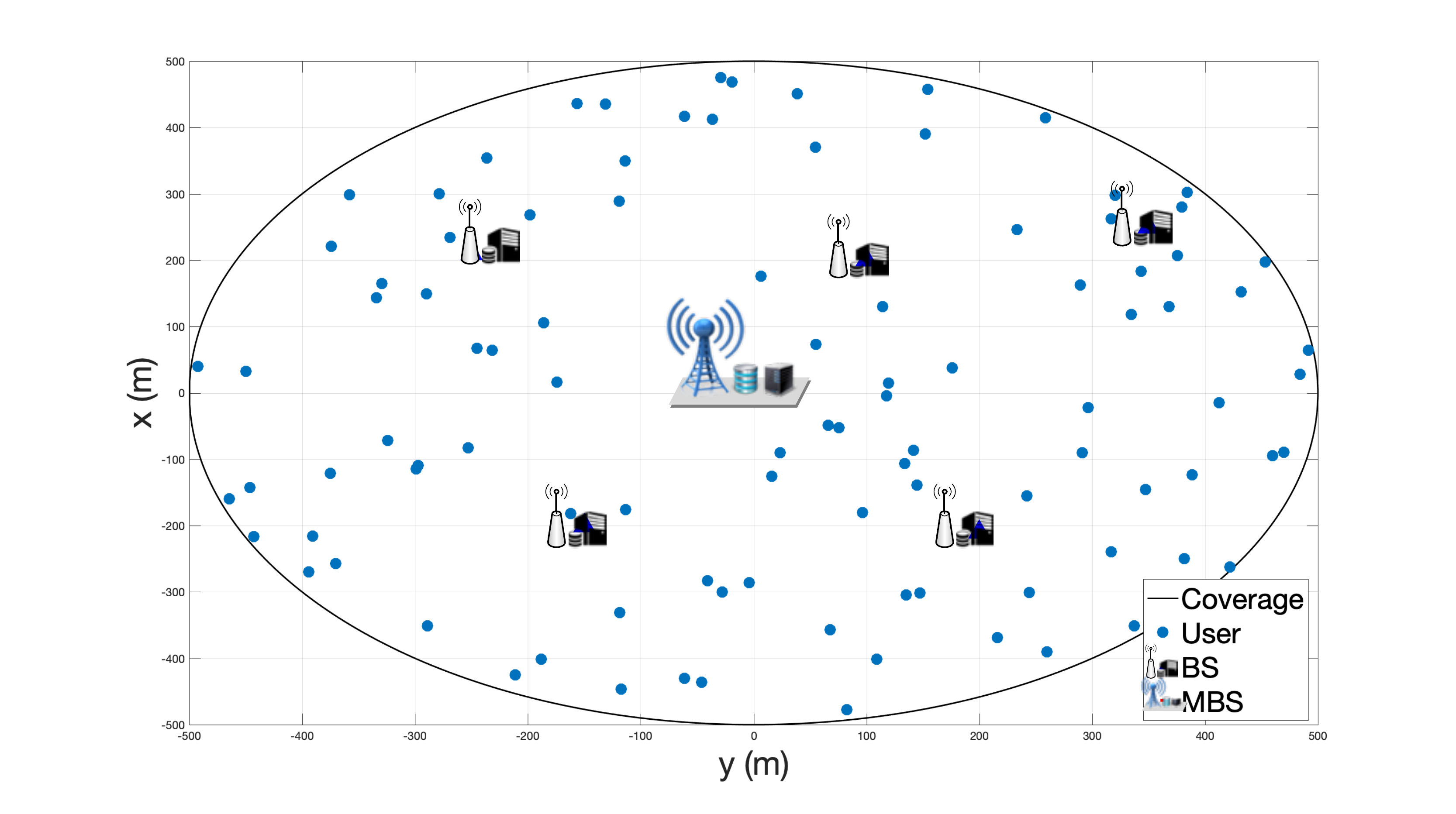}
	\caption{Illustration of the wireless network}
	\label{fig:area}
\end{figure}

We conduct our experiment on a real-world dataset CIFAR10\cite{krizhevsky2009learning}. The CIFAR10 dataset consists of 60,000 images in 10 classes, including 50,000 training images and 10,000 test images. Learning on the image dataset simulates the real edge computing scenarios such as traffic flow monitoring and image recognition of smart cameras. We consider a wireless network with 1 MBS, 5 BSs, and 100 end users. The users are randomly distributed in the coverage area of BSs, as shown in Fig. \ref{fig:area}. The CIFAR10 dataset is shuffled and assigned to the end users randomly. Thus the training data of federated learning is Independent and Identically Distributed (IID) in our experiment. The Convolutional Neural Network (CNN) \cite{cnn} is adopted as the machine learning model for federated learning. The CNN model has two $5 \times 5$ convolution layers (with 32 channels and 64 channels, respectively) and $2 \times 2$ max-pooling layers, followed by a fully connected layer with 512 units.
The maximum CPU frequencies of five BSs are 2.6GHz, 1.8GHz, 3.6GHz, 2.4GHz, and 2.4GHz, respectively. The transmission power of RSUs and the MBS is 34dBm and 42dBm, respectively. The bandwidth of the subchannel is set as 30MHz. $N_0$ is set to -174 dBm. 

\begin{figure}[htb]
	\centering
	\includegraphics[scale=0.18]{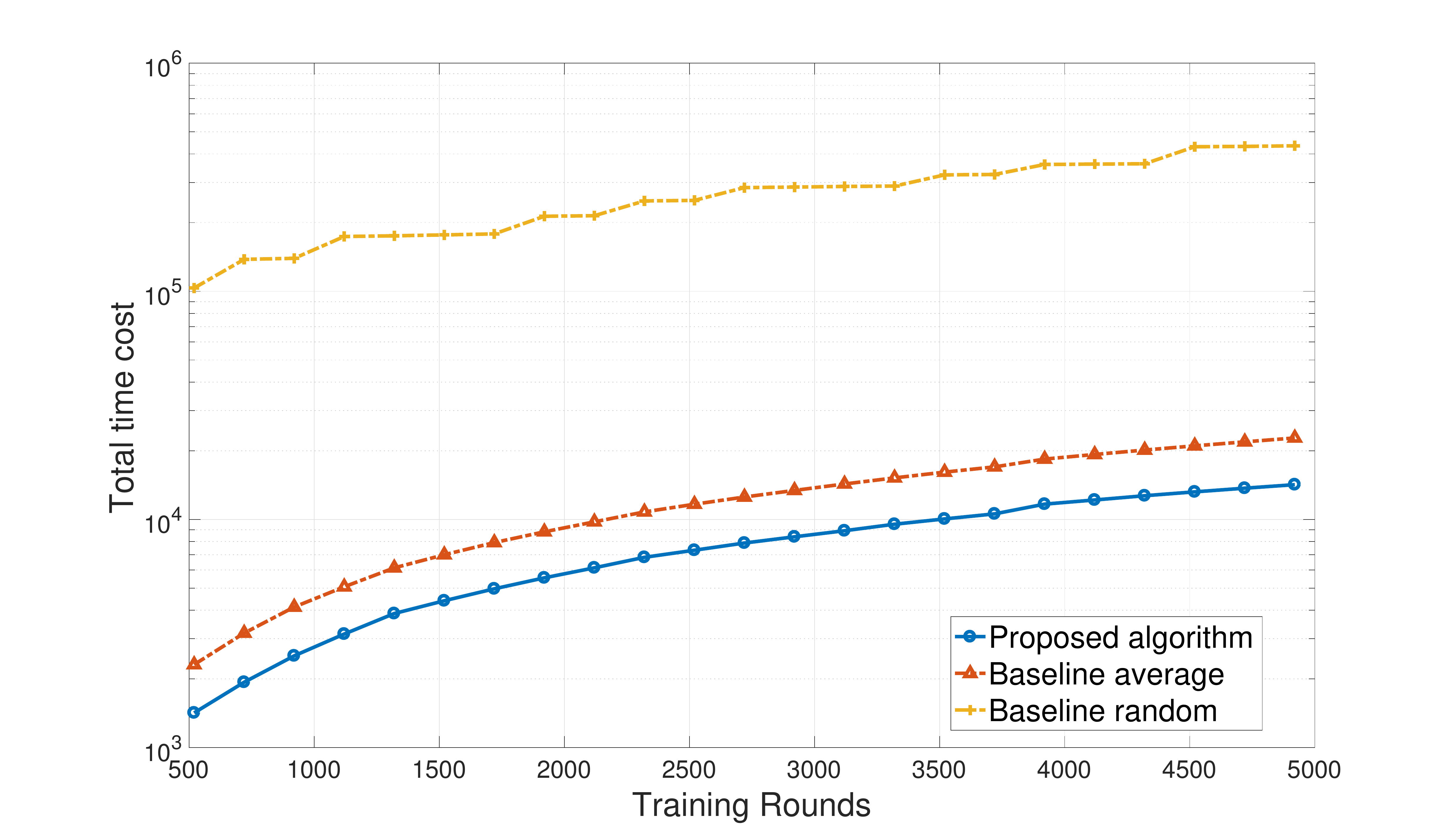}
	\caption{Total system time cost of different schemes in training process}
	\label{fig:FL_time}
\end{figure}

\begin{figure}[htb]
	\centering
	\includegraphics[scale=0.18]{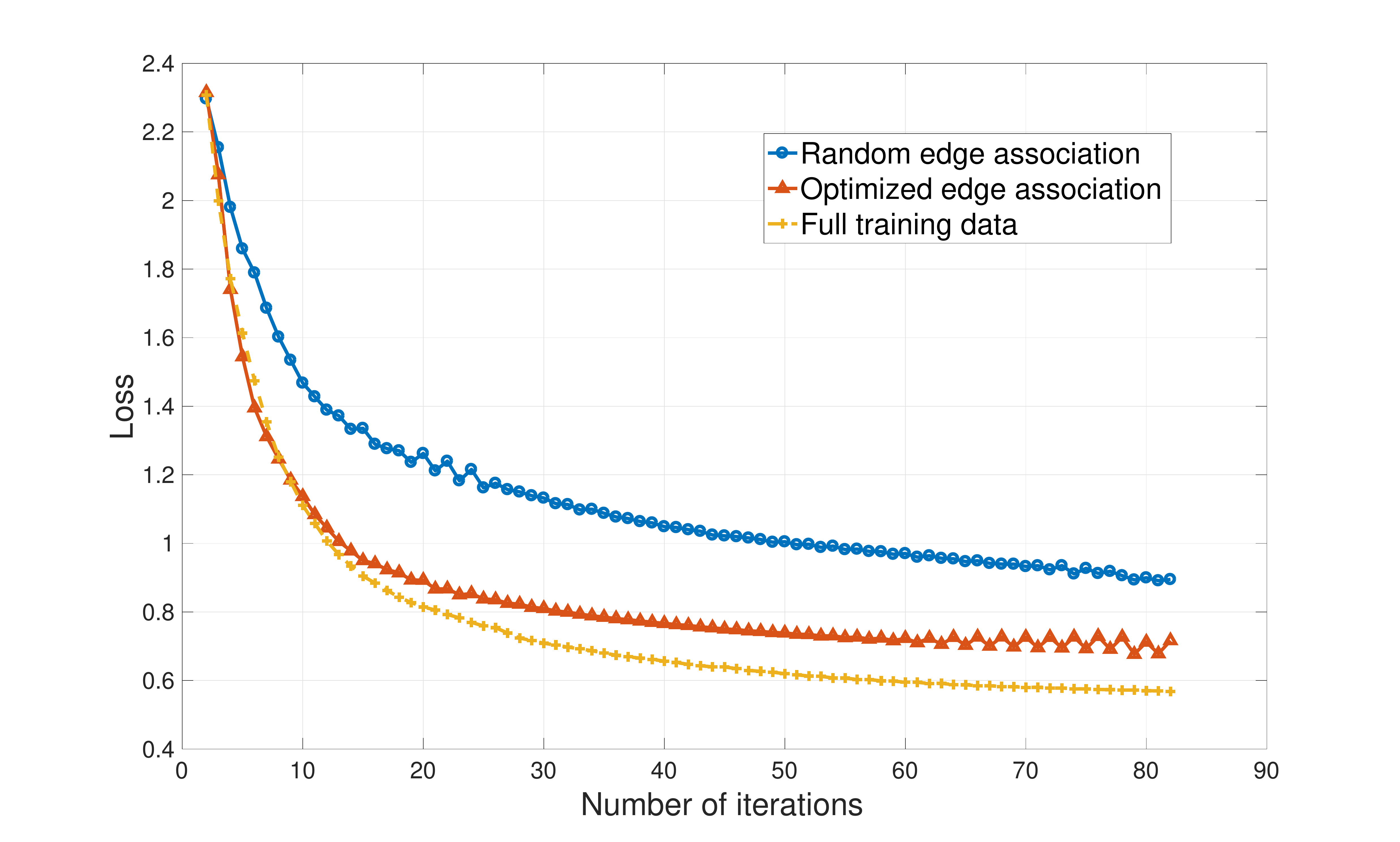}
	\caption{The loss performance in the learning process}
	\label{fig:FL_loss}
\end{figure}

The latency performance with respect to training rounds is shown in Fig. \ref{fig:FL_time} for our proposed algorithm and the conventional learning algorithms with random edge association and with average edge association. Our proposed algorithm significantly reduces the system time cost compared with the benchmark algorithms. The multi-agent reinforcement learning algorithm optimizes digital twin association, and optimally allocates communication resources, which improves the running efficiency and reduces the system latency for our scheme. Fig. \ref{fig:FL_loss} depicts the learning performance of the proposed algorithm, compared with the learning algorithm with full training data and the learning algorithm with random edge association. From the results we can see that our proposed algorithm achieves a comparable convergence performance compared to fully trained federated learning, where all training data of users is utilized in each training process. The learning loss of our proposed algorithm is improved through the optimization process compared with the conventional algorithm with random edge association. 

\begin{figure}[htb]
	\centering
	\includegraphics[scale=0.18]{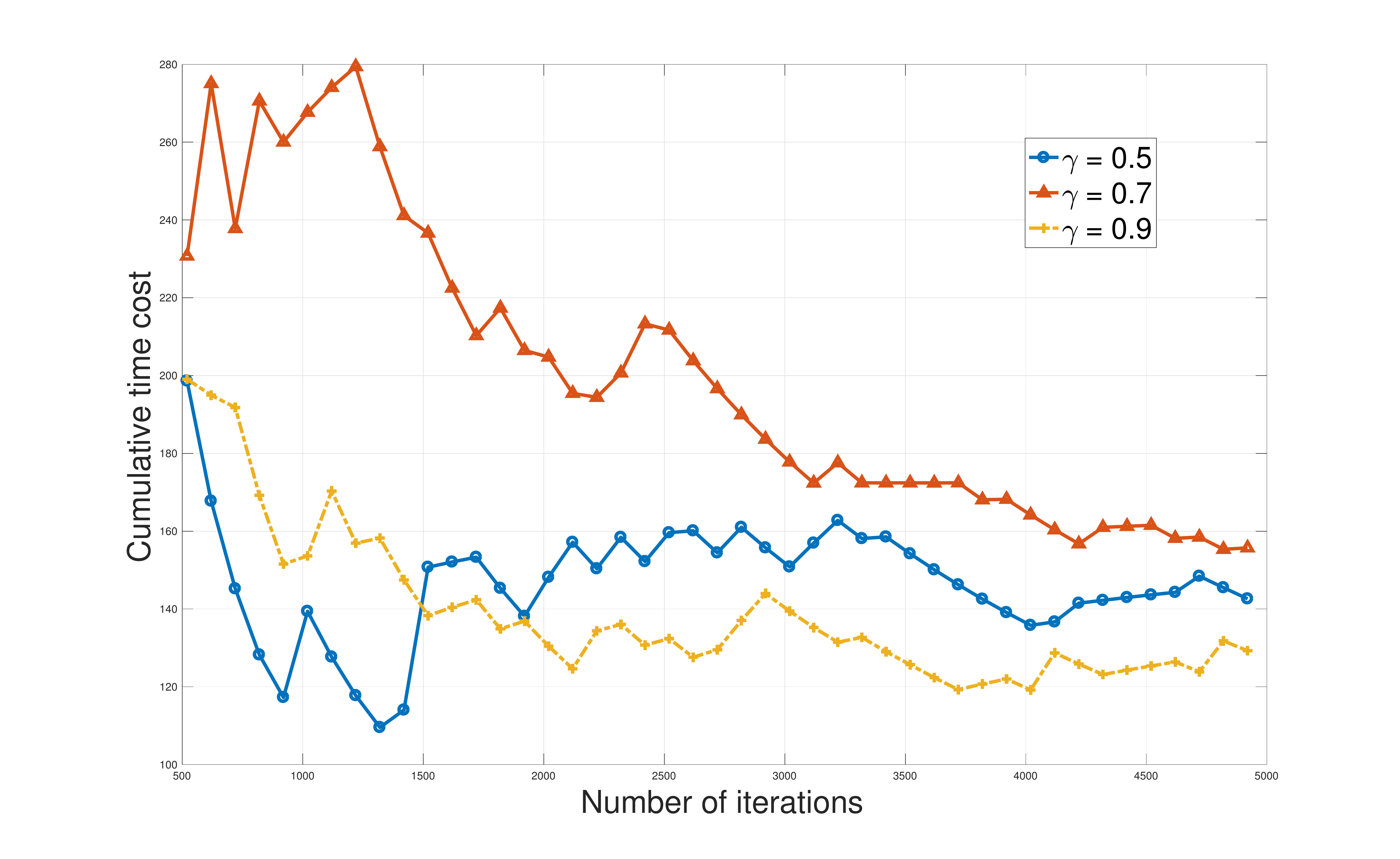}
	\caption{The cumulative average system cost in training process of DRL}
	\label{fig:DRL_cm_reward}
\end{figure}

\begin{figure}[htb]
	\centering
	\includegraphics[scale=0.18]{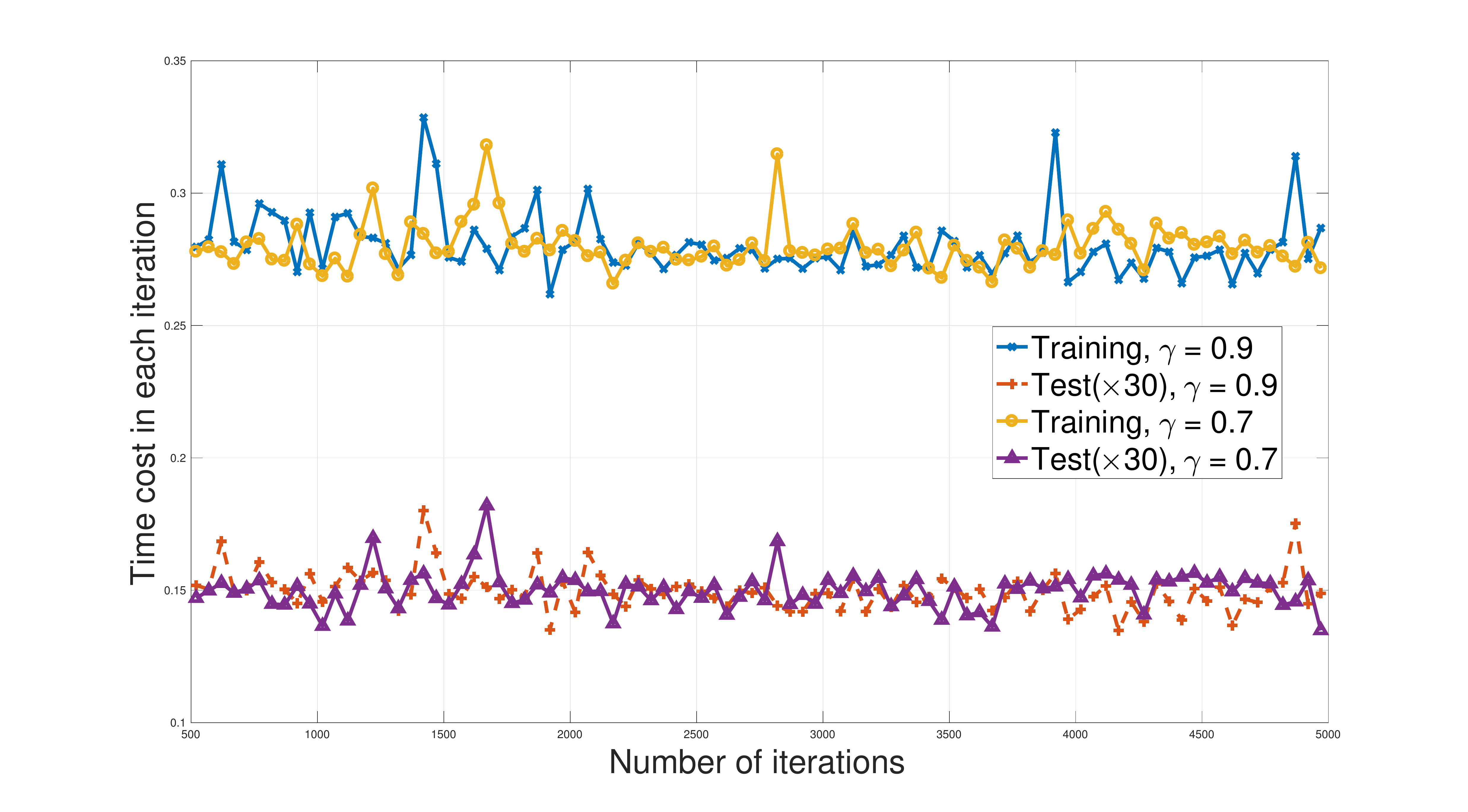}
	\caption{Time cost of the proposed optimization algorithm}
	\label{fig:DRL_time}
\end{figure}

The average cumulative reward is calculated by $\mathcal{R}_n = \frac{\sum_{t=1}^n \sum_{i=1}^M R_{i,t}}{n \cdot M}$, where $R_{i,t}$ is the total reward of agent $i$ in episode $t$, and $M$ is the number of agents. The cumulative average time cost of our proposed multi-agent deep reinforcement learning are depicted in Fig. \ref{fig:DRL_cm_reward} for different values of the discount factor $\gamma$. The cost of different $\gamma$ converges as the number of iterations increases, which denotes that the training process of the DRL model has maximized the cumulative reward and has minimized the system time cost. The curve with $\gamma=0.9$ achieves the best performance, which has a higher convergence rate and smaller system time cost. The time cost for running our reinforcement learning based optimization algorithm is shown in Fig. \ref{fig:DRL_time}. We can see that the training of our proposed optimization algorithm costs much more time than the test phase. The reason is that the training of deep reinforcement learning model is a process of iterative exploration. The agents take much time to interact with the environment to learn optimal policies. Although there are some fluctuations in the result curves, the overall time cost for training or testing in each iteration is comparable. The time is consumed mainly by the training process of the deep neural networks in reinforcement learning. From Fig. \ref{fig:DRL_time} we can also see that a larger discount factor $\gamma$ slightly leads to a larger time cost in each iteration. The reason is that a larger discount factor incurs more computation in the policy training process, thus leading to a larger time cost.

\section{Conclusion}
\label{section:conclusion}
In this paper, we proposed the new DTWN model based on federated learning and blockchain, and improved the running efficiency of the proposed scheme. We first presented the DTWN model in a wireless edge network consisting of end users and BSs. We further developed a blockchain empowered federated learning scheme in DTWN for realizing edge intelligence. To improve the efficiency of the proposed scheme and better allocate the limited resources, we formulated an optimization problem for given learning accuracy. Specifically, we formulated the problem of edge association between digital twins and BSs and derived an optimal solution to the problem by exploiting the multi-agent reinforcement learning. Numerical results based on the real-world dataset demonstrated that the proposed scheme effectively reduces the learning latency and achieves good learning convergence.

% Can use something like this to put references on a page
% by themselves when using endfloat and the captionsoff option.
\ifCLASSOPTIONcaptionsoff
  \newpage
\fi

\bibliography{ref.bib}

\clearpage
\end{document}